\documentclass[preprint,12pt]{elsarticle}
\usepackage{amssymb}
\usepackage{amsmath}
\usepackage{booktabs}
\newcommand{\mycmidrule}[1]{\cmidrule(lr){#1}}
\usepackage{longtable}% for long tables
\usepackage{tabularx}
\usepackage{float}
\usepackage{array}
\usepackage{url} 
\usepackage{hyperref}
\usepackage{multirow}

\usepackage{tcolorbox}
\usepackage[T1]{fontenc}
\usepackage{inconsolata}
\tcbset{
  llmbox2/.style={
    colback=gray!6,
    colframe=gray!50,
    boxrule=0.4pt,
    arc=2pt,
    left=6pt,
    right=6pt,
    top=4pt,
    bottom=4pt
  }
}

\journal{Journal of Biomedical Informatics}

\begin{document}

\begin{frontmatter}

\title{A Concept-Based Interpretable Framework for Clinical Text-based Prediction via Token–Concept Attention without Concept Supervision}

\author[label1]{Jifan Gao \corref{cor}}
\affiliation[label1]{organization={Department of Data Science, Dana-Farber Cancer Institute},
            addressline={3 Blackfan St},
            city={Boston},
            postcode={02115}, 
            state={Massachusetts},
            country={United States}}
\author[label2]{Junjie Hu}\affiliation[label2]{organization={Department of Biostatistics and Medical Informatics, University of Wisconsin-Madison},%Department and Organization
            addressline={610 Walnut St}, 
            city={Madison},
            postcode={53726}, 
            state={Wisconsin},
            country={United States}}
\author[label3]{John Caskey}
\author[label3]{Madeline Oguss}
\author[label1]{Simona Cristea}
\author[label2]{Anoop Mayampurath}
\author[label3]{Matthew Churpek}
\affiliation[label3]{organization={Department of Medicine, University of Wisconsin-Madison},
            addressline={610 Walnut St},
            city={Madison},
            postcode={53726}, 
            state={Wisconsin},
            country={United States}}
\author[label3]{Majid Afshar}
\author[label2]{Guanhua Chen \corref{cor}} %% Author name

\cortext[cor]{Corresponding authors} 
\cortext[email]{Email Addresses: jifan.gao@ds.dfci.harvard.edu (Gao); gchen25@wisc.edu (Chen).}

%% Abstract
\begin{abstract}

Objective: Concept-based interpretation has emerged as a promising approach for explaining blackbox models using human-understandable, high-level concepts. However, most existing methods require annotated concept labels to train the model to identify concepts in the input, and obtaining such annotations in clinical text is often expensive. This study aims to enable concept-based interpretation for language models without concept supervision on clinical prediction tasks.

Methods: We propose a novel framework that enables concept-based interpretation without concept supervision through a tailored cross-attention bottleneck. Clinical text and the text descriptions of medical concepts are first tokenized and embedded by a pretrained language model. A cross-attention module then estimates the presence of each concept for each text token, producing concept relevance scores that capture how concepts are involved throughout the note. These scores are pooled and used in a linear prediction layer to model clinical outcomes in an interpretable manner.

Results: The performance of the proposed framework was compared against blackbox models across two tasks: chest trauma severity stratification using a private dataset and an in-hospital mortality prediction task using the publicly available MIMIC-III dataset. The proposed framework achieved AUROC values within 0.02 of the corresponding blackbox models. The framework also enabled cohort- and patient-level interpretation, capturing the overall concept importance across the cohort  and patient-specific concept contributions.
 
Conclusion: Our proposed framework realizes concept-based interpretation without requiring ground-truth concept presence labels. It improves the transparency of clinical text predictions while maintaining strong performance in clinical prediction tasks.

\end{abstract}

% \begin{center} 
% \includegraphics[width=\textwidth,height=0.75\textheight,keepaspectratio]{diagram.pdf} 
% \end{center} 
% \vspace{0.5em} 
% \noindent 
% \begin{minipage}{\textwidth} 
% \small \textbf{Graphical abstract.} Clinical note tokens and ICD-10 concept descriptions are encoded into a shared embedding space. The token--concept attention bottleneck estimates concept relevance scores, which are then used in a linear prediction layer to support concept-based interpretation without concept supervision. 
% \end{minipage} \end{graphicalabstract}

% \begin{highlights}
% \item We propose a concept-based interpretable framework for clinical language models that does not require concept-level supervision.
% \item A token-concept cross-attention bottleneck enables the final predictions to be formulated as a transparent linear function of concept relevance scores.
% \item Across two different clinical prediction tasks, the framework achieved performance close to blackbox models while providing both cohort- and patient-level explanations.
% \end{highlights}

%% Keywords

\end{frontmatter}



\section{Introduction}
\label{sec:introduction}

Clinical text is largely attributable to medical concepts such as diseases and conditions \cite{navarro2023clinical}, which are the lexicon necessary in clinical decision-making. Clinicians routinely process and incorporate the large vocabulary of medical concepts to accurately assess clinical risks \cite{d2017impact, ivanov2021improving}. Recent advances in clinical language models have enabled strong performance on clinical risk predictions \cite{acharya2024clinical, jiang2023health, redekop2025zero, sakai2026large}. However, these models often encode text into high-dimensional vectors, making it difficult to determine which clinical concepts drive a given prediction. Since medical concepts are explicitly present in clinical narratives and form the basis of the clinical decision process, concept-based interpretable models \cite{kim2018interpretability, koh2020concept, shin2023closer, espinosa2022concept} provide a potential bridge between model predictions and clinically meaningful interpretation \cite{narain2025improving, feng2025bayesian}, supporting transparency in the use of language models in clinical practice.

Concept-based interpretable models usually first map each input to a set of concepts and then use those concepts for prediction \cite{koh2020concept}. However, this concept-mapping step often requires a labor-intensive feature engineering step of concept annotations in addition to outcome labels. Applying concept-based interpretation to clinical text remains challenging because concept annotations require human efforts and additional computational cost. In this study, we develop a framework for making concept-based interpretable predictions from clinical text without requiring concept-level supervision. We use 3-digit ICD-10 codes as the concept vocabulary to build an interpretable bottleneck layer, as ICD-10 provides a clinically applicable and ubiquitous ontology to capture a comprehensive collection of human diseases and conditions \cite{hirsch2016icd}, and ICD-10-based features have been shown to perform well in clinical prediction tasks \cite{gao2025count}. To identify which concepts are present in a clinical note without concept supervision, we design a tailored cross-attention module \cite{rigotti2021attention, bie2024mica} that estimates the relevance of each 3-digit ICD-10 concept to the input text based on semantic similarities between concept text and clinical notes. The final prediction is then generated by a linear function over these identified concepts, allowing the model’s decision process to be modeled by clinically meaningful concepts in an interpretable manner. 

We evaluate the framework on two distinct tasks: chest trauma severity stratification \cite{kulshrestha2021comparison} using a private dataset and in-hospital mortality prediction using MIMIC-III \cite{johnson2016mimic} . Across both tasks, our concept-based model achieves performance comparable to blackbox language models, while offering clear explanations at multiple levels: at the cohort level, it identifies which condition concepts drive predictions across the dataset; at the patient level, it reveals the specific concepts that most influenced a prediction. 

By modeling predictions using meaningful medical concepts rather than opaque latent representations, the outputs of our framework are easier for clinicians to interpret. This can improve interpretation in machine learning-assisted decision-making, help identify potential risk factors, and promote transparency of language models in healthcare settings.

\begin{center} \renewcommand{\arraystretch}{1.25} \makebox[\textwidth][c]{
\begin{tabularx}{1.25\textwidth}{@{}>{\bfseries\raggedright\arraybackslash}p{0.34\textwidth}@{\hspace{1.6em}}X@{}} \toprule \multicolumn{2}{@{}l}{\textbf{Statement of Significance}} \\ 
\midrule

Problem or Issue: &
Existing concept-based interpretable models usually require concept-level annotations, which are expensive to obtain for clinical text. \\[0.6em]
\midrule
What is Already Known: &
Blackbox language models can perform well on clinical prediction tasks, and concept bottleneck models can provide human-understandable explanations when concept labels are available. \\[0.6em]
\midrule
What this Paper Adds: &
We propose a novel framework that enables concept-based interpretation without concept supervision through a tailored cross-attention bottleneck. Across two clinical text prediction tasks, the framework preserved performance close to blackbox models while enabling cohort- and patient-level interpretation.\\[0.6em]
\midrule
Who would benefit from the new knowledge in this paper: &
Biomedical informatics researchers, clinical NLP researchers, and clinicians interested in transparent clinical risk prediction from free-text notes. \\

\bottomrule
\end{tabularx}% 
} \end{center}

\section{Related Work}

Concept-based interpretable models, including concept bottleneck models, typically follow a two-step design: (i) identify a set of concepts involved in each input sample based on a predefined concept set, and (ii) predict the outcome from those identified concepts \cite{koh2020concept}. In practice, the concept identification stage usually requires concept annotations, which introduces an additional labeling step beyond the collection of outcome labels \cite{koh2020concept, shin2023closer, espinosa2022concept, yuksekgonul2023posthoc}. This dependence on concept-level supervision has become one of the main practical challenges to applying concept bottleneck models in real-world settings such as clinical text, where concept annotations are expensive and difficult to obtain.

To reduce the need for manual concept annotation, recent “label-free” variants of concept bottleneck models attempt to replace human concept labels with automatically generated concept annotations. For example, some approaches use large language models to generate or expand concept sets and to produce pseudo-concept labels \cite{oikarinen2023labelfree,tan2023c3m,ludan2023tbm}. These methods reduce the burden of manual labeling, but they still rely on an explicit concept-learning module to train the model to identify which concepts are present in the input \cite{tan2023c3m,ludan2023tbm}. As a result, although the automatic annotation can serve as supervision signals to teach the model to identify concepts from input, the overall pipeline still depends on a separate step that maps the input to concept labels.

Another challenge in concept bottleneck modeling is that the predefined concept set may be incomplete. When important predictive factors are not covered by the selected concepts, predictive performance can deteriorate \cite{liu2025hybrid, bhan-etal-2025-towards}. To compensate for this limitation, some methods introduce an additional residual component that captures predictive signals outside the concept bottleneck \cite{yuksekgonul2023posthoc, tan2024opencbm}. Although this residual pathway can improve predictive performance, it is not concept-based, and therefore, the resulting gain in accuracy comes at the cost of interpretability.

Motivated by these limitations, our work aims to develop a concept-based, interpretable framework for clinical language models that does not require concept annotations and does not rely on an additional uninterpretable residual learning module. Instead of supervising a separate concept predictor, we tailor a token-concept cross-attention bottleneck to estimate concept relevance directly from the semantic space between clinical notes and concept descriptions, and we output predictions through a transparent linear function based on the resulting concept relevance scores.

\section{Methods}
\label{sec:methods}

\begin{figure}[ht!]
\centering
\includegraphics[width=\textwidth]{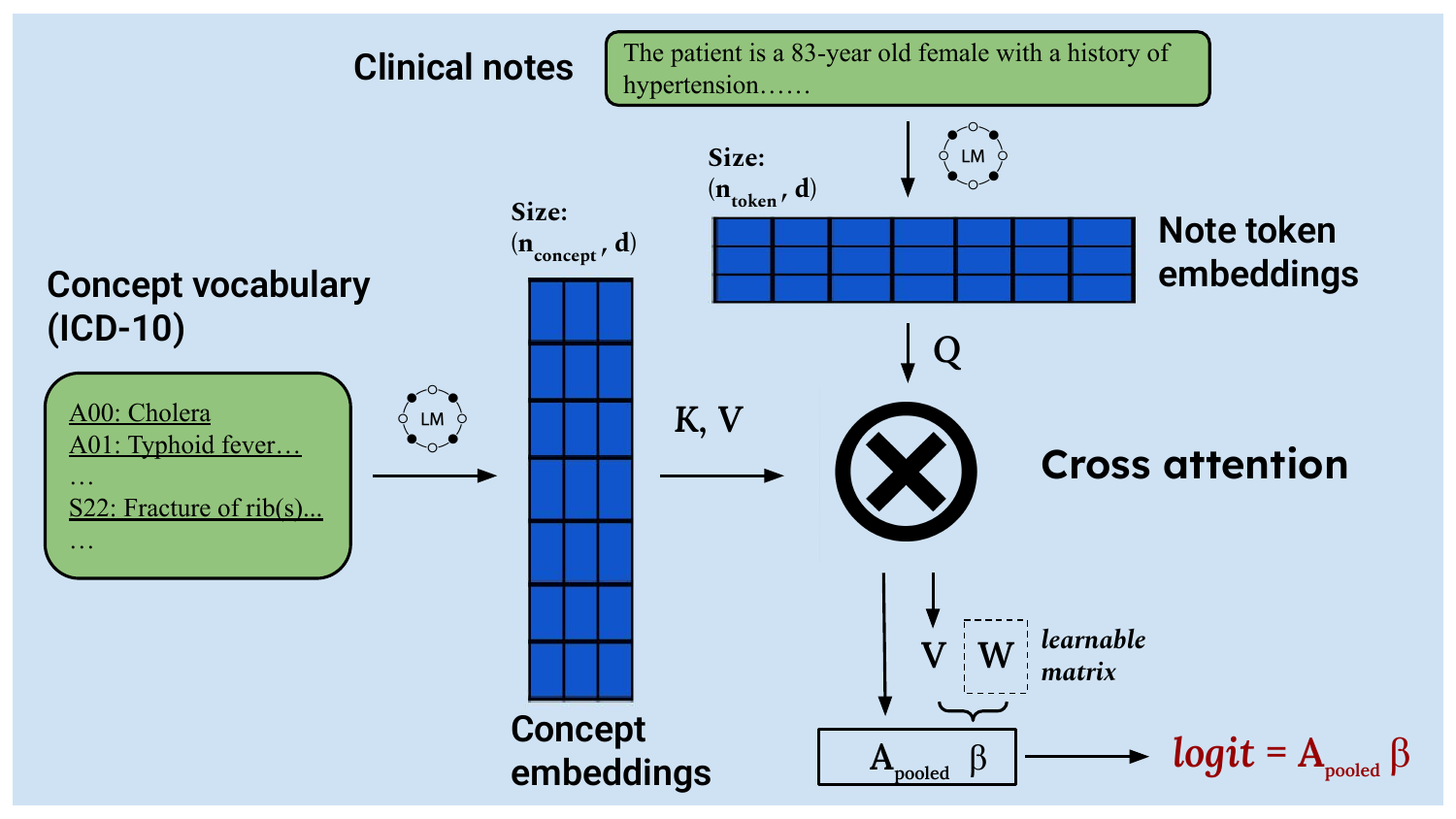}
\caption{Methodological Diagram: Clinical note tokens and concepts (3-digit ICD-10 codes) are encoded into a shared embedding space using the same pretrained language model. The token-concept cross-attention bottleneck estimates how strongly each concept is reflected in the note, and the pooled concept scores are used as covariates in a linear function to calculate the outcome logit.}
\label{fig:diagram}
\end{figure}

Figure \ref{fig:diagram} shows the architecture of our proposed design. We first tokenize clinical text and encode these tokens into embedding vectors using a pretrained language model. Using the same language model, we embed the text description of each concept in the concept vocabulary (e.g., 3-digit ICD-10 codes) and pool the resulting embeddings into a single vector representation for each concept. A token-concept cross-attention module then models the alignment between note token embeddings and concept embeddings, estimating how strongly each concept is present in the input text. These token-level concept relevance scores are subsequently pooled into note-level concept relevance scores, which are used as covariates in a linear prediction layer to estimate the outcome logit in an interpretable manner.

\subsection{Setup}

Let $\mathcal{D}=\{(x_i, y_i)\}_{i=1}^N$ denote a dataset of clinical notes and outcomes.
Each note $x_i$ is a sequence of tokens $x_i=(w_{i,1},\dots,w_{i,L})$ (padded/truncated to a maximum length $L$),
and $y_i$ is the target outcome.
We also have a fixed concept vocabulary
$\mathcal{C}=\{c_1,\dots,c_M\}$ (e.g., 3-digit ICD-10 concepts), where each concept $c_j$ is associated with a text description
$d_j$.

To model clinical text tokens like stopwords that do not correspond to any concept, we introduce a \emph{null concept} $c_0$
with a non-natural sentinel description $d_0$ (e.g., \texttt{<<NULL\_CONCEPT>>}). More details about the construction and implementation of the \emph{null concept} are presented in Supplementary Materials B.
The indices of concepts range in $j\in\{0,1,\dots,M\}$, where $j=0$ denotes the null concept.

\subsection{Text encoder for free text and concept descriptions}
\label{method:encoder}

\paragraph{Text and concept embedding}
We use a pretrained language model as backbone encoder $f_{\theta}(\cdot)$ that maps a sequence of note tokens to contextual token embeddings.
For note $x_i$, the encoder outputs
\begin{equation}
\mathbf{H}_i = f_{\theta}(x_i) \in \mathbb{R}^{L \times d},
\end{equation}
where $\mathbf{H}_i[p]\in\mathbb{R}^d$ is the $d$-dimensional embedding of token position $p$.

To place concepts and note tokens in a shared embedding space, we encode each concept description using the \emph{same} language model. For each concept text $d_j$, we extract a single embedding vector with a dimension of $d$ by applying a pooling operation to the encoder outputs, such as max pooling, mean pooling, or taking the embedding of the \texttt{[CLS]} token when available:
\begin{equation}
\mathbf{e}_j = \mathrm{Pool}_{p\in\{1,\dots,L\}} f_{\theta}(d_j) \in \mathbb{R}^{d},
\end{equation}
where $\mathbf{e}_j $ denotes the pooled embedding of the concept description. We take the embedding of the \texttt{[CLS]} token as the default pooling operation. We stack all concept embeddings into a matrix
\begin{equation}
\mathbf{E} = 
\begin{bmatrix}
\mathbf{e}_0^\top\\
\mathbf{e}_1^\top\\
\vdots\\
\mathbf{e}_M^\top
\end{bmatrix}
\in \mathbb{R}^{(M+1)\times d}.
\end{equation}
Throughout, we treat $\mathbf{H}_i$ as text embeddings and $\mathbf{E}$ as concept embeddings.

\paragraph{Warmup and freezing}
We conduct a short task-adaptive warmup to improve the alignment between note-token embeddings and concept embeddings for the prediction task. Specifically, we first fine-tune the encoder $f_{\theta}(\cdot)$ for the target prediction task. For each $x_i$, we obtain pooled representations by using max pooling, mean pooling, or the embedding of the \texttt{[CLS]} token when available. The embedding of the \texttt{[CLS]} token is used as the default pooling operation. We then attach feedforward layers on top of this pooled patient embedding to predict the outcome. After warmup, we compute $\mathbf{H}_i$ and $\mathbf{E}$ using the warmed-up encoder, and finally freeze $f_{\theta}(\cdot)$ when training the concept-based predictor (Section~\ref{sec:concept_model}).

\subsection{Token--concept cross-attention bottleneck}
\label{sec:concept_model}

A cross-attention bottleneck is tailored to quantify how each concept is reflected in every note token. The note token embeddings \(\mathbf{H}_i\) are passed through fully learned linear projection layers as the query matrix \(\mathbf{Q}_i \in \mathbb{R}^{L \times d}\), while the concept embeddings are passed through separate fully learned linear projection layers as the key matrix \(\mathbf{K} \in \mathbb{R}^{(M+1)\times d}\) and the value matrix \(\mathbf{V} \in \mathbb{R}^{(M+1)\times d}\).

For each token \(p\), we first compute cosine similarities with all real
concepts. Let \(\mathcal{T}_{i,p}\) denote the indices of the top \(k\)
real concepts with the highest cosine similarities. The attention weights
are then computed as
\begin{equation}
\begin{aligned}
s_{i,p,j}
&= \cos\!\left(\mathbf{Q}_i[p], \mathbf{K}[j]\right),
\qquad j=1,\ldots,M, \\[4pt]
\ell^{\mathrm{null}}_{i,p}
&=
\frac{m_{\mathrm{null}}
-\max_{r=1,\ldots,M}s_{i,p,r}}
{\tau_{\mathrm{null}}}
+b_{\mathrm{null}}, \\[4pt]
\tilde{s}_{i,p,j}
&=
\begin{cases}
s_{i,p,j}/\tau_{\mathrm{attn}},
    & j\in\mathcal{T}_{i,p},\\
\ell^{\mathrm{null}}_{i,p},
    & j=0,\\
-\infty,
    & \text{otherwise},
\end{cases} \\[4pt]
\mathbf{A}_{i,p,:}
&=
\operatorname{sparsemax}
\left(\tilde{\mathbf{s}}_{i,p,:}\right).
\end{aligned}
\end{equation}
Here, \(j=0\) denotes the NULL concept, \(k=4\), and
\(\mathcal{T}_{i,p}=\operatorname{TopK}_k(s_{i,p,1:M})\).
Thus, \(A_{i,p,j}\) quantifies the attention assigned by token \(p\)
in note \(i\) to concept \(j\).

To obtain note-level concept covariates, we first pool the token-level attention weights across all token positions. Specifically, for note \(i\), the pooled attention assigned to concept \(j\) is defined as
\begin{equation}
\alpha_{i,j} = \mathrm{Pool}_{p\in\{1,\dots,L\}} A_{i,p,j},
\end{equation}
where \(A_{i,p,j}\) denotes the attention weight between token \(p\) and concept \(j\). Thus, \(\alpha_{i,j}\) serves as a note-level quantification of the extent to which concept \(j\) is involved in the note. We use max pooling as the default pooling option to capture the relevance of each concept.

The final prediction is then formulated as a linear function with $\alpha_{i,j}$ as the concept covariates. The product of the concept value matrix \(\mathbf{V} \in \mathbb{R}^{(M+1)\times d}\) and a learnable projection \(\mathbf{W} \in \mathbb{R}^{d \times K}\) acts as the coefficient matrix.
\begin{equation}
\boldsymbol{\beta} = \mathbf{V}\mathbf{W} \in \mathbb{R}^{(M+1)\times K},
\end{equation}
where \(\beta_{j,k}\) quantifies the contribution of concept \(j\) to class \(k\). The output logits for class $k$ are therefore given by

\begin{equation}
\hat{z}_{i,k} = \sum_{j=0}^{M} \alpha_{i,j}\beta_{j,k} + b_k.
\end{equation}

This design makes the prediction depend explicitly on token-concept attention and supports a transparent linear concept-level decision rule. The model output is directly interpretable: the pooled attention weights \(\alpha_{i,j}\) act as covariates, while \(\mathbf{V}\mathbf{W}\) provides the corresponding coefficients, yielding a linear model over concept relevance scores. We apply \(L_1\) regularization to \(\beta_{j,k}\), encouraging only a small subset of concepts to have nonzero contributions for each class. We use a weighted loss during training to handle class imbalance. All hyperparameters are selected using a tuning subset split from the development set, and the default values are reported in the Supplementary Materials C.

\subsection{Interpretation}
\label{sec:interpretation}
\paragraph{Cohort-level interpretation}
\label{subsec:cohort_interpretation}
Across the dataset, we estimate which concepts are most influential by ranking concepts using their weight magnitude $\|\beta_{j,:}\|$.

\paragraph{Patient-level interpretation}
For a specific note $x_i$, we quantify the contribution of each concept $j$ to class $k$ as
$\alpha_{i,j}\beta_{j,k}$ and identify which tokens $p$ support a specific concept $j$ by inspecting the token-level attention scores
$A_{i,p,j}$ (e.g., highlighting tokens with the largest $A_{i,p,j}$ for top-contributing concepts).

\subsection{Concept Perturbation Analysis}

We validate whether the concepts identified by our design can capture important signals when using $f_{\theta}$ as a blackbox model. For each test sample $i$ and class $k$, we rank each concept $j$ by patient-specific contribution $\alpha_{i,j}\beta_{j,k}$ and select the top 5 or 10 concepts as the top contributors for that class. Tokens having non-zero attention scores for these concepts are masked by \texttt{[MASK]} token. We constructed a token-count-matched random control for each positive- and negative-class perturbation: for every note, we randomly selected the same number of non-padding tokens as in the corresponding contributing concept perturbation and replaced them with \texttt{[MASK]}. As an additional robustness analysis, we replaced the concept-associated tokens with randomly sampled tokens from the model vocabulary. These perturbed test notes are then passed to the corresponding blackbox model, and the resulting change in predictive performance is used to assess whether the top-contributing concepts identified by our framework reflect signals utilized by the blackbox model.

\subsection{Concept grounding validation}

To assess whether contributing concepts identified by the framework were supported by the notes, we conducted a grounding validation focused on the most cohort-level contributing concept described in Methods \ref{subsec:cohort_interpretation} for each backbone across both prediction tasks. For each concept, we used DeepSeek Llama3 Distilled \cite{deepseekai2025deepseekr1incentivizingreasoningcapability} to verify whether it was present in the notes in which the framework identified it as the top contributor. To validate the LLM-based assessments, we randomly sampled 60 note-concept pairs and had two reviewers with biomedical informatics backgrounds assess concept presence. Both the LLM and human reviewers were instructed to count a concept as present only when it was explicitly documented. Mentions that were negated, historical, or uncertain were not considered present. We then used the validated LLM procedure to extend the analysis to all notes associated with the top concepts.

\section{Results}

\subsection{Cohort}

We evaluated our design on real-world clinical prediction tasks using two different datasets. The first task focused on stratifying chest trauma injury severity using a private dataset from the UW Health System, with binary labels indicating whether a patient has serious chest trauma injuries defined by the Abbreviated Injury Scale \cite{kulshrestha2021comparison, gennarelli2006ais} annotated by human experts. Clinical text inputs included ED notes and radiology reports documented within the first 8 hours after admission, following the approach in \cite{gao2024automated}. Eligible notes were ordered by documentation time after admission. The second task used the publicly available MIMIC-III database \cite{johnson2016mimic} to predict in-hospital mortality for ICU admissions as a binary outcome. We followed the cohort and admission-note construction procedures described by \cite{van2021clinical}, using documentation available at admission while excluding sections describing the subsequent hospital course or discharge, as well as direct indications of mortality. For both datasets, each patient had one encounter. Input sequences exceeding the maximum context length of the backbone encoder were truncated to the first 512 tokens after note construction.

We conducted temporal validation for the chest trauma severity stratification task. The development set used data collected from January 2015 to December 2018, and the test set used data collected from January 2019 to December 2019. Because MIMIC-III has dates shifted for PHI protection, we randomly picked 80\% of the encounters as the development set and the remaining 20\% for testing following \cite{van2021clinical}. For both tasks, we used a concept library composed of 3-digit ICD-10 codes \cite{hirsch2016icd}, totaling 2,347 unique concepts. The cohort characteristics are shown in Table \ref{tab:cohort_char}

\begin{table}[ht]
\centering
\caption{Cohort characteristics and label distributions}
\resizebox{\textwidth}{!}{
\begin{tabular}{*5l}
\toprule
{}   & \multicolumn{2}{c}{Chest trauma severity stratification} & \multicolumn{2}{c}{In-hospital mortality prediction}  \\
{}   &  Development &  Test &  Development &  Test \\
\midrule
Encounters, n & 7,693 & 1,983 & 36,803 & 9,291 \\
Age, median (IQR) &  59 (38,75) & 60 (44,77) & 65 (52,77) & 65 (52,77)\\
Female, n (\%) & 3,225 (41.9) & 855 (43.1) & 16,078 (43.7) & 4,078 (43.9)\\
White, n (\%) & 7,096 (92.2) & 1,802 (90.9) & 26,449 (71.9) & 6,635 (71.4)\\
\midrule
\multicolumn{5}{l}{Number of pre-truncation tokens by backbone} \\
\quad\quad SapBERT/ClinicalBERT & 406 (353,431) & 396 (336,425) & 477 (286,702) & 476 (284,710)\\
\quad\quad BGE & 437 (379,464) & 431 (362,462) & 527 (319,772) & 525 (318,780)\\
\midrule
\multicolumn{5}{l}{Labels for chest trauma severity stratification} \\
\quad\quad Non-serious & 5,695 (74.0) & 1,482 (74.7) & - & -\\
\quad\quad Serious or greater & 1,998 (26.0) & 501 (25.3) & - & -\\
\midrule
\multicolumn{5}{l}{Labels for in-hospital mortality} \\
\quad\quad Negative & - & - & 34,839 (94.7) & 8,807 (94.8)\\
\quad\quad Positive &- & - & 1,964 (5.3) & 484 (5.2)\\
\bottomrule
\end{tabular}
}
\label{tab:cohort_char}
\end{table}

\subsection{Predictive performance}

As encoder-only architectures have been serving as strong defaults in clinical classification tasks \cite{sakai2026large, gao2024automated, tinn2023fine, gruenloh2025integrating}, we evaluated our concept-based interpretable design using three encoder-based backbones as $f_{\theta}$: \textsc{SapBERT}, a BERT-family \cite{devlin2019bert} model trained to align biomedical entity representations; \textsc{ClinicalBERT} \cite{alsentzer2019publicly}, a clinical domain-adaptation of BERT trained on clinical notes; and \textsc{bge-large-en-v1.5} \cite{bge_embedding}, a more recent general embedding model with strong performance on large-scale text embedding benchmarks \cite{myers2025lessons}. For each backbone, we compared the concept-based interpretable model with its corresponding blackbox model for chest trauma severity prediction and in-hospital mortality prediction. 

We also implemented the standard concept bottleneck framework for each backbone as a baseline. In particular, we first used MedCat \cite{kraljevic2019medcat} to annotate ICD-10 concepts in the notes, as MedCat's annotations have been reported to help achieve stronger clinical NLP performance compared to using other tools \cite{gao2023hybrid} like MetaMap \cite{aronson2010overview}. Each backbone was then trained to predict these concept labels, and the resulting concept predictions were used to predict the outcome using a logistic regression. 

Quantitative evaluations are shown in Table \ref{tab:primary_results}. Overall, our framework preserved predictive performance well compared with the original blackbox models, especially when using SapBERT and ClinicalBERT as backbones, with performance remaining aligned across both tasks. Although the \textsc{bge-large-en-v1.5} showed a slight reduction, the overall results indicate that our framework can maintain performance comparable to their underlying blackbox models. As shown in Table \ref{tab:supp_warmup_ablation}, removing task-adaptive warmup substantially reduced performance across all backbones and both tasks .

\newcommand{\best}[1]{{\bfseries\boldmath #1}}
\newcommand{\nearbb}{\ensuremath{^{\dagger}}}

\begin{table}[ht]
\centering
\caption{\textbf{Quantitative performance on the test set across three random seeds.} Quantitative performance (mean $\pm$ std over three random seeds) on the test set for chest trauma severity prediction and in-hospital mortality prediction. For each pretrained language model, we compared the original black-box backbone, a standard concept bottleneck model (CBM), and our proposed framework. The best-performing model within each model group and metric is bolded, and $^{\dagger}$ is present when performance is within 0.02 difference of the corresponding black-box model. Across \textsc{SapBERT}, \textsc{ClinicalBERT}, and \textsc{bge-large-en-v1.5}, the proposed framework achieved performance close to that of the corresponding blackbox model while substantially outperforming the standard CBM baseline.}
\resizebox{\textwidth}{!}{
\begin{tabular}{*5l}
\toprule
{} & \multicolumn{2}{c}{Chest trauma severity} & \multicolumn{2}{c}{In-hospital mortality} \\
{} & AUROC & AUPR & AUROC & AUPR \\
\midrule
SapBERT blackbox & \best{$0.962\,{\scriptstyle \pm\,0.012}$} & \best{$0.873\,{\scriptstyle \pm\,0.010}$} & \best{$0.845\,{\scriptstyle \pm\,0.017}$} & \best{$0.346\,{\scriptstyle \pm\,0.015}$} \\
SapBERT + Standard CBM & $0.917\,{\scriptstyle \pm\,0.023}$ & $0.741\,{\scriptstyle \pm\,0.014}$ & $0.728\,{\scriptstyle \pm\,0.030}$ & $0.209\,{\scriptstyle \pm\,0.019}$ \\
Ours (SapBERT backbone) & $0.959\,{\scriptstyle \pm\,0.013}$\nearbb & \best{$0.873\,{\scriptstyle \pm\,0.009}$}\nearbb & $0.839\,{\scriptstyle \pm\,0.016}$\nearbb & $0.332\,{\scriptstyle \pm\,0.018}$\nearbb \\
\midrule
ClinicalBERT & \best{$0.959\,{\scriptstyle \pm\,0.011}$} & \best{$0.871\,{\scriptstyle \pm\,0.010}$} & \best{$0.849\,{\scriptstyle \pm\,0.020}$} & \best{$0.325\,{\scriptstyle \pm\,0.017}$} \\
ClinicalBERT + Standard CBM & $0.881\,{\scriptstyle \pm\,0.034}$ & $0.702\,{\scriptstyle \pm\,0.025}$ & $0.664\,{\scriptstyle \pm\,0.045}$ & $0.153\,{\scriptstyle \pm\,0.027}$ \\
Ours (ClinicalBERT backbone) & $0.957\,{\scriptstyle \pm\,0.012}$\nearbb & $0.864\,{\scriptstyle \pm\,0.011}$\nearbb & $0.833\,{\scriptstyle \pm\,0.022}$\nearbb & $0.314\,{\scriptstyle \pm\,0.019}$\nearbb \\
\midrule
BGE & \best{$0.957\,{\scriptstyle \pm\,0.010}$} & \best{$0.861\,{\scriptstyle \pm\,0.012}$} & $0.839\,{\scriptstyle \pm\,0.024}$ & $0.309\,{\scriptstyle \pm\,0.015}$ \\
BGE + Standard CBM & $0.757\,{\scriptstyle \pm\,0.041}$ & $0.640\,{\scriptstyle \pm\,0.039}$ & $0.673\,{\scriptstyle \pm\,0.051}$ & $0.158\,{\scriptstyle \pm\,0.033}$ \\
Ours (BGE backbone) & $0.937\,{\scriptstyle \pm\,0.014}$\nearbb & $0.821\,{\scriptstyle \pm\,0.013}$ & \best{$0.843\,{\scriptstyle \pm\,0.016}$}\nearbb & \best{$0.318\,{\scriptstyle \pm\,0.018}$}\nearbb \\
\bottomrule
\end{tabular}
}
\label{tab:primary_results}
\end{table}

\subsection{Cohort-level interpretation}
\label{subsec:global_interpretation}

We present cohort-level concept-based interpretation in Table \ref{tab:global_interpretation_backbones}, showing which medical concepts have the highest impacts on the model’s predictions across the study cohort. Within each backbone, the top three contributing concepts were consistent across three random seeds. For chest trauma severity, the SapBERT- and ClinicalBERT-based models highlighted concepts that were consistent with the task, including fractures involving the ribs or thoracic region and pneumothorax-related conditions. The BGE-based model identified concepts that were, in general, related to the injury region, but the selected concepts were less consistent compared with SapBERT and ClinicalBERT, and included  non-trauma-related conditions. For in-hospital mortality, the top-ranking concepts were less aligned across backbones, indicating that concept attribution across the dataset for this task is more sensitive to the choice of backbones. Our additional grounding analysis in Table \ref{tab:concept_grounding} further showed that SapBERT, which was pretrained to align synonymous biomedical concepts, achieved stronger concept grounding across both tasks. These findings indicate that careful backbone selection is important for reliable concept-level interpretation.

\begin{table}[H]
\centering
\caption{\textbf{Cohort-level concept interpretation across backbone encoders.} For chest trauma severity, the highest-ranked concepts were broadly aligned with clinically expected injury patterns, especially for SapBERT- and ClinicalBERT-based models. In contrast, the highest-ranked concepts for in-hospital mortality were more variable across backbones, reflecting varied behaviors when using different backbones.}
\label{tab:global_interpretation_backbones}
\renewcommand{\arraystretch}{1.15}
\setlength{\tabcolsep}{5pt}
\begin{tabularx}{\textwidth}{>{\raggedright\arraybackslash}p{2.5cm} >{\raggedright\arraybackslash}X >{\raggedright\arraybackslash}X >{\raggedright\arraybackslash}X}
\toprule
\textbf{Task} & \textbf{SapBERT-Backbone} & \textbf{ClinicalBERT-Backbone} & \textbf{BGE-Backbone} \\
\midrule
\textbf{Chest trauma severity}
& 1) Rib, sternal, and thoracic spine fracture\newline
  2) Pyothorax\newline
  3) Pneumothorax
& 1) Rib, sternal, and thoracic spine fracture\newline
  2) Pyothorax\newline
  3) Pneumothorax
& 1) Pleural plaque\newline
  2) Pleural effusion\newline
  3) Pneumothorax
\\
\midrule
\textbf{In-hospital mortality}
& 1) Hypotension\newline
  2) Hepatic failure, not elsewhere specified\newline
  3) Pleural effusion in conditions classified elsewhere
& 1) Cardiac arrest\newline
  2) Adverse effects, not elsewhere specified\newline
  3) Respiratory tuberculosis
& 1) Tuberculosis of nervous system\newline
  2) Heart Failure\newline
  3) Respiratory failure
\\
\bottomrule
\end{tabularx}
\end{table}

\subsection{Patient-level interpretation}
\label{subsec:local_interpretation}

\begin{figure}[H]
\centering
\includegraphics[width=0.9\textwidth]{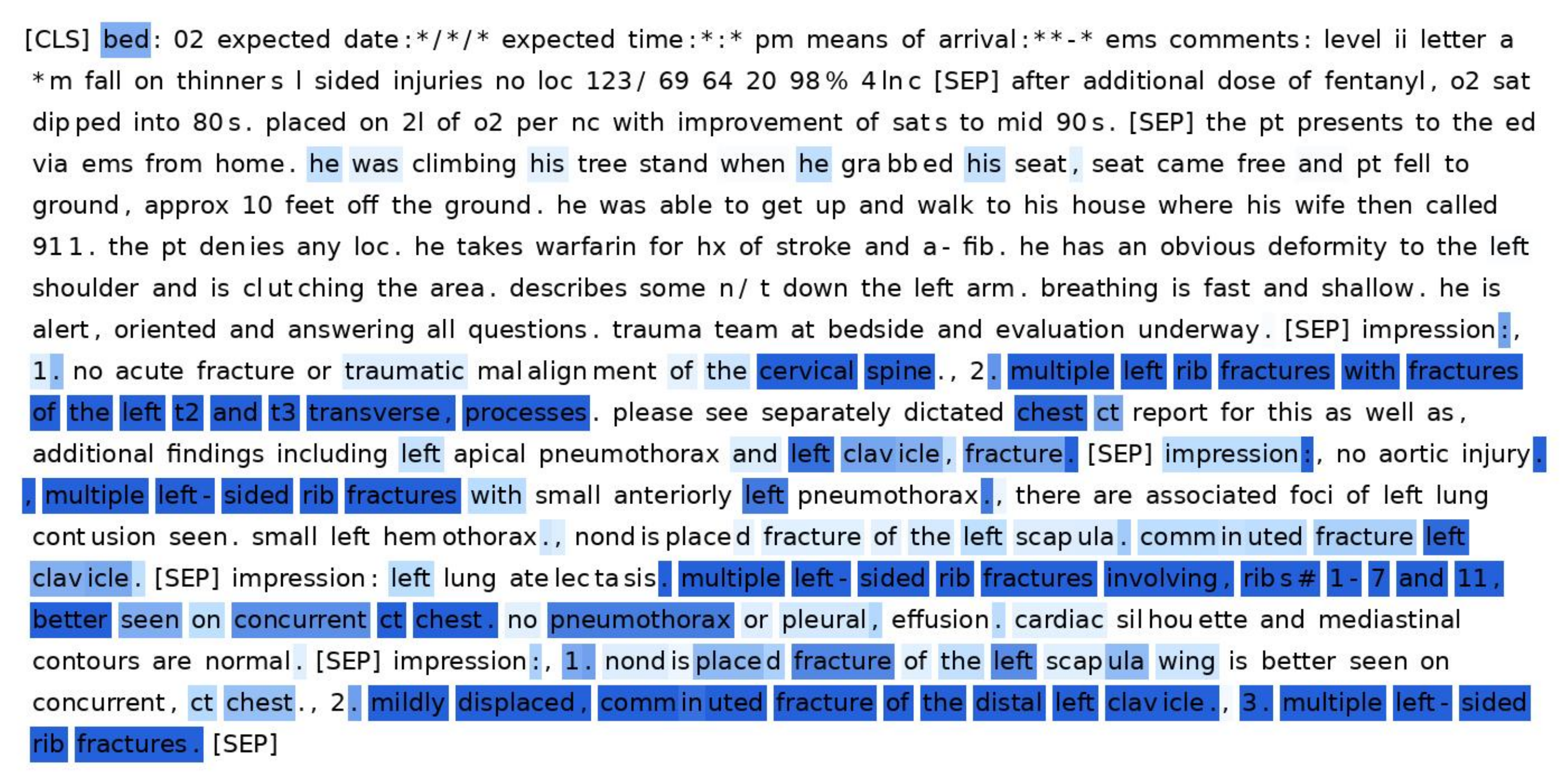}
\caption{\textbf{Patient-level interpretation examples for chest trauma severity prediction.} Using the interpretation procedure described in Section \ref{sec:interpretation}, the framework with \textsc{SapBERT} as the backbone encoder identified rib fractures as the top contributing concept in the chest trauma severity example. The darkness of the highlighted tokens indicates the attention weight assigned to the corresponding top concept, with darker blue representing attention weights closer to 1 and lighter blue representing attention weights closer to 0. }
\label{fig:local_interpretation}
\end{figure}

Figure \ref{fig:local_interpretation} presents examples of patient-level interpretations produced by the proposed framework using SapBERT as the backbone for the chest trauma severity task. Following the interpretation procedure described in Section \ref{sec:interpretation}, the framework identified rib fractures as the top contributing concept in the chest trauma severity example. The darker the blue, the higher the attention score for the top contributing concepts (rib fractures) in the examples. In the chest trauma severity task, the model assigns strong attention to text spans describing rib fractures, which aligns well with the clinical severity of thoracic injury. The patient-level interpretation of other backbones in chest trauma severity task is provided in the Supplementary Materials A. Mortality prediction examples are not shown in accordance with the MIMIC data use policies.

As an illustrative comparison to LLM generation, we queried DeepSeek Llama3 Distilled \cite{deepseekai2025deepseekr1incentivizingreasoningcapability} on the same chest trauma and compared its reasoning with the patient-level interpretation produced by our design. A summarized comparison between the patient-level interpretation of our framework (\textsc{SapBERT} as the backbone encoder) and the LLM output is shown in Figure \ref{fig:ds_output}. The full prompt and output are presented in the Supplementary Materials D. In the chest trauma example, the LLM identified multiple rib fractures as the top factor indicating severe chest trauma, consistent with our framework’s identification of rib fractures as the top contributing concept. 

\begin{figure}[H]
\centering
\includegraphics[width=\textwidth]{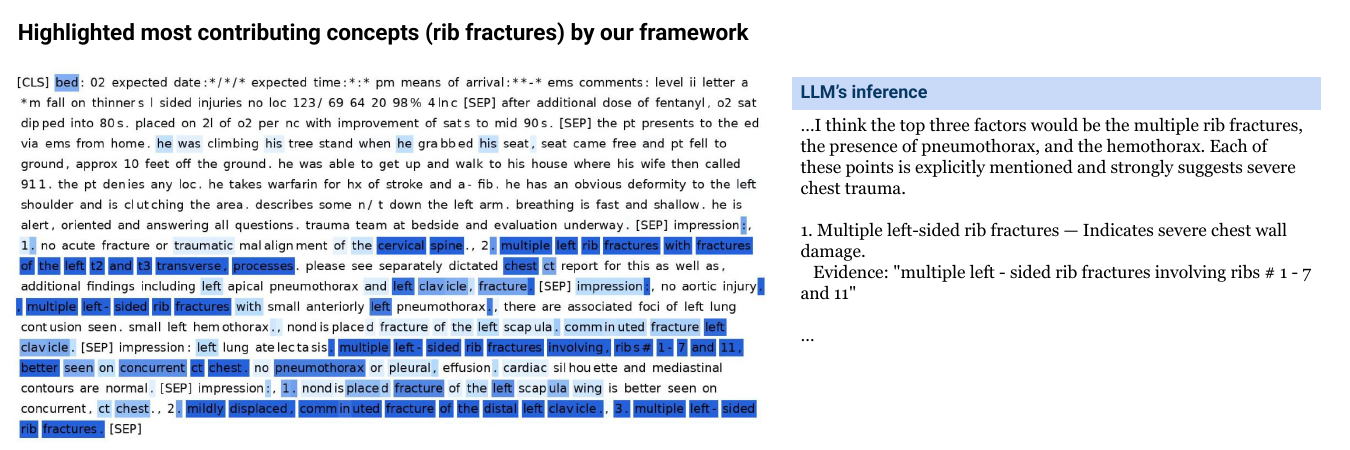}
\caption{\textbf{Comparison of patient-level interpretations between the proposed framework and LLM-generated reasoning.} In the chest trauma example, the LLM identified multiple rib fractures as the top factor indicating severe chest trauma, which was consistent with the framework’s identification of rib fractures as the top contributing concept.}
\label{fig:ds_output}
\end{figure}

\subsection{Blackbox Model Behavior under Concept Perturbation}

\begin{figure}[H]
\centering
\includegraphics[width=\textwidth]{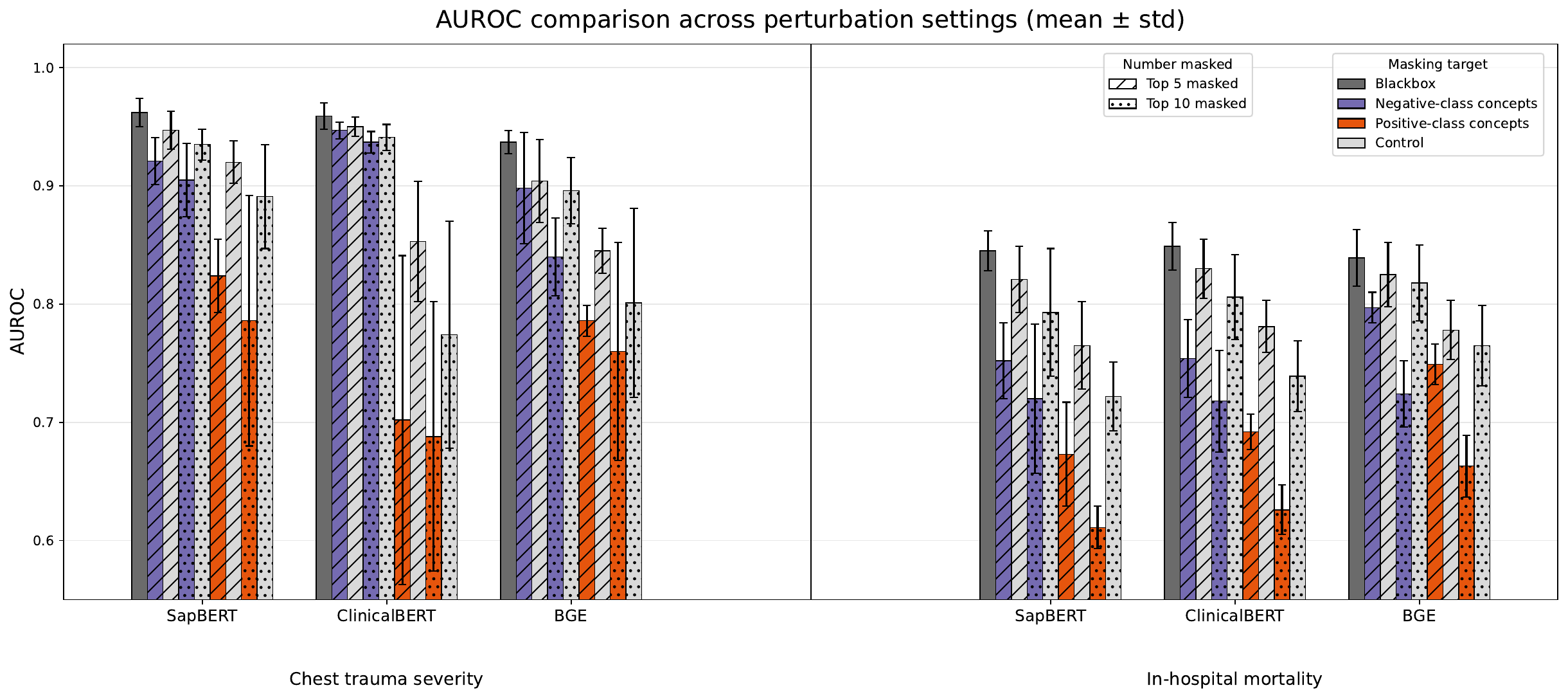}
\caption{\textbf{Perturbation study evaluating whether concepts identified by our framework reflect predictive signals used by the blackbox models.} For each blackbox backbone (\textsc{SapBERT}, \textsc{ClinicalBERT}, and \textsc{bge-large-en-v1.5}) and each task (chest trauma severity and in-hospital mortality), we compared AUROC after masking tokens. Across backbones and tasks, masking top-contributing positive- or negative-class concepts consistently produced larger AUROC drops than token-count-matched random perturbation. Masking positive-class contributors led to larger performance drops. Masking the top 10 class-contributors resulted in a larger performance drop than masking the top 5.}
\label{fig:perturbation}
\end{figure}

Figure \ref{fig:perturbation} shows the perturbation analysis, where we measured AUROC after masking or replacing tokens associated with concepts identified by our framework as having large class-specific contributions. Numerical results are presented in Supplementary Table \ref{tab:supp_perturbation_control} and \ref{tab:supp_perturbation_control_random_tokens}. Number and percentage of tokens being masked are shown in Table \ref{tab:supp_masked_tokens}. Across all backbones and both tasks, masking or replacing concepts identified as top contributors to either the positive or negative class led to larger AUROC drops than token-count-matched random perturbation. This pattern suggests that the top contributing concepts identified by our framework reflect signals that the backbone model leverages for prediction. Also, masking positive-class contributors led to performance declines, indicating that these ``positive'' concepts may carry strong signals for supporting the target prediction. In addition, the performance drop was greater when masking the top 10 contributing concepts than when masking the top 5, indicating that removing a larger set of contributing concepts has a stronger impact on model behavior.

\subsection{Concept grounding validation}

With the top concepts reported in Table~\ref{tab:global_interpretation_backbones} for each task, inter-reviewer agreement was 96.7\%, and samples with agreed human judgments showed approximately 94.8\% agreement with the LLM assessments. We extended the validated LLM procedure to assess concept grounding across all corresponding notes. As shown in  Table~\ref{tab:concept_grounding}, the model with the SapBERT backbone demonstrated strong concept grounding precision across both tasks: rib fracture was confirmed in 294/418 (70.3\%) chest trauma notes in which it was identified as a top contributor, and hypotension was confirmed in 38/49 (77.6\%) mortality notes. In comparison, ClinicalBERT and BGE backbones showed lower and more variable grounding precision across the two tasks. 

\begin{table}[htbp]
\centering
\caption{Validation of attention-derived concept grounding. For each task and backbone,
we selected the top concept identified by the framework and evaluated all notes in which
that concept was highlighted by attention scores. Concept presence was independently
assessed by DeepSeek Llama3 Distilled based on the corresponding clinical note.
“Our highlighted” denotes the number of notes in which the concept was identified as the
top contributor, while “LLM confirmed” denotes the number of those notes for which the
LLM also considered the concept to be present.}
\label{tab:concept_grounding}

\small
\begin{tabularx}{\linewidth}{
    @{}
    l
    l
    >{\raggedright\arraybackslash}X
    r
    @{}
}
\toprule
Task & Backbone & Concept &
\shortstack{LLM confirmed /\\Our highlighted (\%)} \\
\midrule
Mortality
    & SapBERT
    & Hypotension
    & 38/49 (77.6) \\

    & ClinicalBERT
    & Cardiac arrest
    & 2/21 (9.5) \\

    & BGE
    & Nervous tuberculosis
    & 48/344 (14.0) \\
\midrule
Chest Trauma
    & SapBERT
    & Rib fracture
    & 294/418 (70.3) \\

    & ClinicalBERT
    & Rib fracture
    & 351/643 (54.6) \\

    & BGE
    & Pleural plaque
    & 251/464 (54.1) \\
\bottomrule
\end{tabularx}
\end{table}

\section{Discussion}

In this study, we propose a framework that enables concept-based interpretation of clinical text without requiring concept labels. We validated the proposed framework on two separate tasks. The results show that the proposed framework can achieve performance comparable to that of blackbox models, while providing cohort- and patient-level interpretation that cannot be directly achieved by a blackbox model.

A key advantage of this design is that prediction is produced by a supervised classifier, with the output explicitly explained through human understandable clinical concepts. Although LLMs have shown promising capabilities in clinical summarization, reasoning, and generation, they do not directly provide task-specific probabilistic risk estimations without additional engineering \cite{shoham2024clinical,gu2024probabilistic,gao2025moma}. Our framework bridges this gap by preserving the supervised classifier as the source of prediction while adding concept-based interpretability to the same language model, allowing the model to provide probabilistic estimations, while providing concept-based explanations that are valuable in clinical applications.

The core component is the cross-attention bottleneck, which automatically estimates concept relevance rather than relying on a supervised concept-learning stage. Given embedding representations of note tokens and candidate concept descriptions, the cross-attention module estimates how strongly each token is associated with each concept, thereby producing concept relevance scores directly from the input text. Since prior work has reported that attention weights may not provide faithful explanations of model behavior \cite{jain2019attention}, we do not use the raw attention weights as explanations. Instead, interpretation in our framework is tied to the model's actual decision function: for each class, the contribution of each concept is computed as the product of its pooled concept relevance score and its class-specific coefficient. The output logit is therefore a sum of concept-level contributions for each sample, so the reported cohort- and patient-level explanations directly reflect the decision process.

Related concept-bottleneck style designs have been explored previously in computer vision, but they are typically implemented in two different ways. In one formulation \cite{bie2024mica}, concepts are used as keys, allowing the hidden representations of the sample input to remain in the value matrix. This design can preserve richer contextual information and may therefore benefit predictive performance, but the resulting representation is less compatible with a transparent linear formulation in which the final prediction is expressed directly in terms of concept relevance scores. In the other formulation \cite{rigotti2021attention}, concepts are used as the values and keys, and the note tokens are used as queries, so that the outcome can be modeled explicitly as a linear function of concept-level covariates. This makes the prediction more transparent and easier to interpret, though it may result in a predictive performance drop because the final step no longer directly leverages the full representation of sample-specific input. Our framework builds on the second manner because it is better aligned with our goal of making clinical text predictions directly explained by human-understandable medical concepts.

As shown in Table \ref{tab:primary_results}, our framework achieves comparable performance to that of corresponding blackbox backbone models. One possible reason that predictive performance did not decline substantially is that our framework uses the complete set of 3-digit ICD-10 codes as candidate covariates for the final predictions. Existing studies have shown that ICD-10–based representations can achieve strong performance across various clinical prediction tasks \cite{tran2022icd, placido2023deep, shmatko2025learning}. By leveraging a comprehensive disease vocabulary rather than a narrowly selected concept set, the model is less likely to miss important risk factors. This may have mitigated the information loss that often occurs in concept bottleneck models when the concept set is incomplete.

We introduce a warm-up stage in which the backbone encoder is first trained on the prediction task using pooled hidden states before applying the concept-based cross-attention module. The goal of this stage is to encourage the encoder to learn task-relevant representations in advance, so that the concept-based model begins from a more informative representation space. The comparative results with and without warm-up training are provided in the Supplementary Materials F.

For in-hospital mortality prediction, the highest-ranked concepts showed greater variability across backbone encoders than in the chest trauma severity task. This suggests that cohort-level concept attribution for mortality is affected more by the representation learned by the backbone encoder. A likely reason is that in-hospital mortality is more heterogeneous. Unlike the chest trauma severity stratification task, which is more closely tied to a set of injury-related concepts, mortality may reflect diverse signals from physiologic measurements, medication exposure, and disease severity. Some of these signals may not be fully captured by the ICD-10 concept vocabulary, leading different backbones to emphasize different concept patterns.

We examine the framework across three different blackbox backbone encoders: \textsc{SapBERT}, \textsc{ClinicalBERT}, and \textsc{bge-large-en-v1.5}. \textsc{SapBERT} and \textsc{ClinicalBERT} represent well-established BERT-based encoders adapted to biomedical and clinical language, respectively. We included \textsc{bge-large-en-v1.5} to test whether a newer, high-performing embedding model could also serve as an effective backbone for the proposed interpretable design. The results suggest that the framework is not tied to a single encoder family, but can be implemented on top of multiple pretrained text encoders while maintaining its interpretability advantages.

Our grounding analysis revealed that concept grounding performance varied across backbone models. The model with the SapBERT backbone demonstrated strong grounding precision across both prediction tasks with LLM-based confirmation, with rib fracture and hypotension in 70.3\% and 77.6\% of notes in which they were identified as top contributors. Grounding performance, however, shows variations across backbone models. SapBERT is a domain-adapted biomedical encoder pretrained to align synonymous medical concepts through contrastive learning, making its representation space suited to the token-concept matching used in our framework. This suggests that careful selection of the backbone model is important for achieving well-grounded concept-based interpretations in our proposed framework.

Task-adaptive warmup plays an important role in maintaining predictive performance in our framework by adapting the encoder representation toward the downstream outcome, as reflected by the substantial performance decrease without warmup (Supplementary Table \ref{tab:supp_warmup_ablation}). Importantly, the adaptation does not remove semantic grounding: the same warmed encoder represents both clinical text and concept descriptions, and our grounding validation showed that the resulting representations, particularly with SapBERT, remained well aligned with clinically meaningful concepts. These findings suggest that warmup provides an effective task-relevant initialization while preserving useful concept-level structure.  

This study has the following limitations. First, our evaluation focused on classification tasks, where the final prediction is binary and can be naturally expressed as a function of concept relevance scores. Second, inputs were truncated to a maximum of 512 tokens, which may exclude information from longer clinical records. Therefore, the patient-level interpretations produced by our framework should be recognized as explanations of the text available to the model, rather than of the complete clinical note. Third, decoder-only models are not included in our experiments as they are more naturally suited to generation tasks. Future work should therefore examine whether the proposed framework can be extended to generation-oriented clinical applications and decoder-based architectures on long clinical text.

\section{Conclusion}

This work presents a concept-based, interpretable framework for clinical text that does not require concept annotations and can be implemented on top of various pretrained encoder backbones. By combining a cross-attention bottleneck with a linear concept-based prediction layer, the framework enables both cohort- and patient-level interpretation while preserving predictive performance close to that of black-box models. Together, these findings suggest that clinically meaningful and more transparent language model predictions can be achieved without additional concept annotations or significant drops in performance, providing a practical direction for interpretable clinical text prediction.

\section*{Acknowledgments}
This was supported in part by the National Institutes of Health/National Institute on Drug Abuse R01DA051464 (MA, MO), National Library of Medicine R01LM012973 (MA), the National Science Foundation DMS-2515263 (GC),  1R01GM169395-01 National Institute of General Medical Sciences (GC), and the Patient-Centered Outcomes Research Institute (PCORI) Award ME-2024C1-37433 (JG, GC). The statements in this work are solely the responsibility of the authors and do not necessarily represent the views of the Patient-Centered Outcomes Research Institute (PCORI), its Board of Governors, or the Methodology Committee.

\section*{Declaration of generative AI and AI-assisted Technologies}

During the preparation of this work, the author(s) used AI tools in order to polish the writing.

\section*{Declaration of Competing Interest}

Simona Cristea is currently employed by Tempus AI, Boston, United States, but this work was conducted before this employment, and the company had no role in the study. The remaining authors declare that they have no known competing financial interests or personal relationships that could have appeared to influence the work reported in this paper.

\section*{Ethical Approval and Informed Consent}

The chest trauma severity stratification component of this study was reviewed and approved by the University of Wisconsin--Madison Minimal Risk Research Institutional Review Board under protocol \#2019-1258, with a waiver of informed consent. The MIMIC-III in-hospital mortality prediction component used a publicly available, de-identified dataset accessed through PhysioNet.

\section*{Data Availability}

The MIMIC-III dataset used for the in-hospital mortality prediction task is available through PhysioNet, subject to PhysioNet's data access requirements. The chest trauma severity stratification data were obtained from the UW Health system after approval from the Institutional Review Board. Our data use agreements do not permit sharing clinical data. Researchers with an interest in accessing the data can reach out to the corresponding authors or Madeline Oguss at \href{mailto:mkoguss@medicine.wisc.edu}{mkoguss@medicine.wisc.edu}. The chest trauma example used for demonstration has been reviewed and approved by the UW HIPAA Privacy Officer. Mortality prediction examples are not shown in accordance with the MIMIC data use policies.

\section*{Code Availability}
The code for MIMIC-III mortality prediction task is available at \url{https://github.com/GGGGFan/concept_attention_without_concept_label}. The chest trauma severity stratification dataset cannot be publicly released due to legal and regulatory constraints.

% \appendix
% \section{Appendix Section}

%\clearpage
\bibliographystyle{elsarticle-num} 
\bibliography{reference.bib}

\appendix
\renewcommand{\thefigure}{S.\arabic{figure}}
\setcounter{figure}{0}
\renewcommand{\thetable}{S.\arabic{table}}
\setcounter{table}{0}

\clearpage
\section*{Supplementary Materials}

\subsection*{A. Patient-level interpretation with ClinicalBERT and BGE backbones}

\begin{figure}[ht!]
\centering
\includegraphics[width=\textwidth]{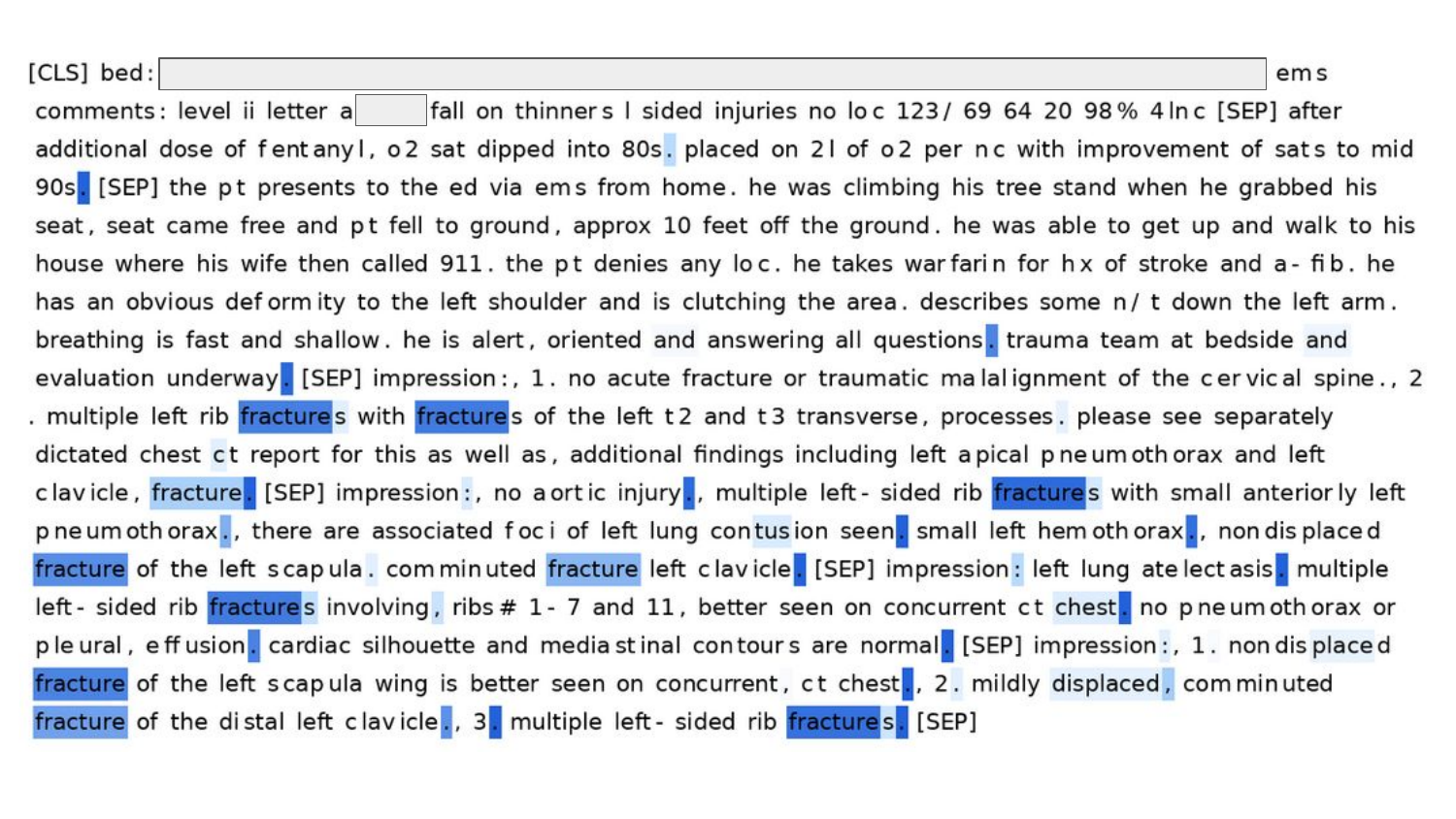}
\vspace{-0.5in}
\caption{ClinicalBERT as the backbone encoder. Rib fractures are identified as the top contributing concept
in the chest trauma severity example. }
\label{fig:appendix_interpretation_clinicalBERT}
\end{figure}

\begin{figure}[ht!]
\centering
\includegraphics[width=\textwidth]{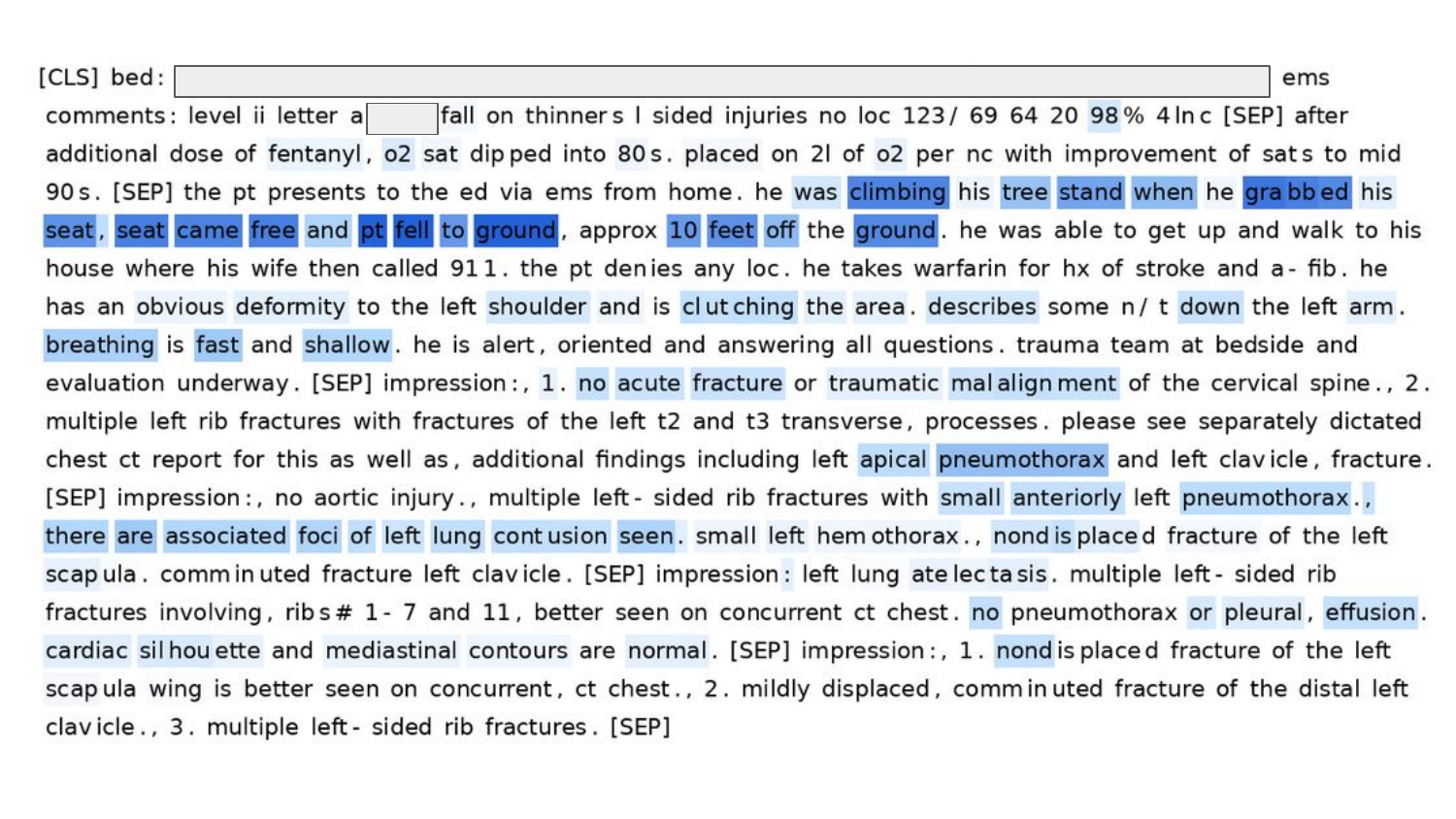}
\caption{BGE as the backbone encoder. Rib fractures are identified as the top contributing concept
in the chest trauma severity example. }
\vspace{-1in}
\label{fig:appendix_interpretation_BGE}
\end{figure}

The patient-level interpretation when using ClinicalBERT and BGE as the backbone is shown in Figure \ref{fig:appendix_interpretation_clinicalBERT} and \ref{fig:appendix_interpretation_BGE}. Both the ClinicalBERT and BGE backbones identify rib fractures as the top contributing concept. ClinicalBERT assigns higher attention weights to explicit mentions of rib fractures, whereas BGE shows more dispersed attention across the relevant clinical context.

\clearpage

\subsection*{B. Additional implementation details for null-concept attention}

To implement the null concept described in the main Methods, we added a null-attention logit for each token before normalizing the token--concept attention scores. Let \(s_{p,j}\) denote the cosine similarity between the embedding of token \(p\) and the embedding of real concept \(c_j\), \(j \in \{1,\ldots,M\}\). For computational sparsity, only the top \(k=4\) real concepts with the largest \(s_{p,j}\) values were retained for each token, and all remaining real-concept logits were masked before attention normalization. The retained real-concept logits were scaled by an attention temperature \(\tau_{\mathrm{attn}}=0.07\). The null-concept logit was defined as
\[
\ell_{p,0}
=
\frac{m_{\mathrm{null}} - \max_{j \geq 1} s_{p,j}}{\tau_{\mathrm{null}}}
+
b_{\mathrm{null}},
\]
where \(m_{\mathrm{null}}=0.85\) is the null-gate margin, \(\tau_{\mathrm{null}}=0.05\) is the null-gate temperature, and \(b_{\mathrm{null}}\) was initialized to 2.0. This design assigns higher null-concept attention to tokens whose maximum similarity to any real ICD-10 concept is low, which is consistent with the role of \(c_0\) as a placeholder for stopwords or other tokens that do not correspond to medical concepts. Attention weights over the null concept and the retained real concepts were normalized using sparsemax. A small entropy penalty with weight \(\lambda_{\mathrm{entropy}}=0.02\) was also applied to the conditional distribution over real concepts to discourage widespread assignment across many concepts. 

\clearpage

\subsection*{C. Hyperparameters}

\newcolumntype{Y}{>{\raggedright\arraybackslash}X}

\begin{longtable}{p{3.0cm} p{4.5cm} p{7.2cm}}
\caption{Default hyperparameters.}
\label{tab:hyperparameters_concept_attention} \\
\toprule
\textbf{Category} & \textbf{Hyperparameter} & \textbf{Value / setting} \\
\midrule
\endfirsthead

\toprule
\textbf{Category} & \textbf{Hyperparameter} & \textbf{Value / setting} \\
\midrule
\endhead

\midrule
\multicolumn{3}{r}{Continued on next page} \\
\endfoot

\bottomrule
\endlastfoot

Concept vocabulary
& Concept source
& Three-digit ICD-10 concept vocabulary. \\

& Concept pooling
& \texttt{[CLS]} pooling. \\

& Concept maximum length
& 256 tokens. \\

& Concept embedding batch size
& 4. \\

\midrule
Backbone encoder
& Pretrained model
& \texttt{SapBERT}. \\

& Input maximum length
& 512 tokens. \\

& Batch size
& 4. \\

& Number of workers
& 2. \\

\midrule
Blackbox warmup model
& Warmup architecture
& Encoder \texttt{[CLS]} embedding followed by a linear classification head. \\

& Number of output classes
& 2. \\

& Warmup optimizer
& AdamW. \\

& Warmup learning rate
& \(1.0 \times 10^{-6}\). \\

& Warmup epochs
& 9. \\

& Warmup loss
& Cross-entropy loss. \\

\midrule
Concept-attention model
& Concept value dimension \(d\)
& 256. \\

& Text embedding freezing
& \texttt{True}. \\

& Concept embedding freezing
& \texttt{True}. \\

& Attention activation
& Sparsemax. \\

& Top-\(k\) concept restriction per token
& 4. \\

& Attention temperature
& 0.07. \\

& NULL gate margin
& 0.85. \\

& NULL gate temperature
& 0.05. \\

& NULL bias initialization
& 2.0. \\

& Note-level concept pooling
& Max pooling over token-level attention scores. \\

\midrule
Concept-model training
& Optimizer
& AdamW. \\

& Learning rate
& \(5.0 \times 10^{-7}\). \\

& Epochs
& 1. \\

& Prediction loss
& Cross-entropy loss. \\

& Entropy penalty weight
& \(\lambda_{\mathrm{entropy}} = 0.02\). \\

& $L_{1}$ regularization weight
& \(\lambda_{\mathrm{1}} = 0.02\). \\

\midrule
Reproducibility
& Random seed
& 42. \\

& Deterministic training
& Enabled. \\

& Tokenizer parallelism
& Disabled. \\

\end{longtable}

\subsection*{D. Prompt and full LLM output on patient-level interpretation examples in \ref{subsec:local_interpretation}
}

\begin{tcolorbox}[llmbox2,title=\textbf{Prompt}]
{\ttfamily\small
You are a medical assistant.

Task:
From the free-text note only, identify the top factors suggesting severe chest trauma.

Output rules:
- Return ONLY the top 3-5 factors, ranked.
- Keep the whole response under 150 words.
- Use only information explicitly stated in the note.
- Do not mention mortality, prognosis, or treatment advice.
- Do not invent vitals, labs, or injuries not written in the note.

For each factor, use exactly this format:
1. [factor] — [why it suggests severe chest trauma, <=12 words]
   Evidence: "[short quote <=8 words]"

If the note is insufficient, say:
"Insufficient evidence for severe chest trauma."

No introduction. No conclusion. No extra explanation.
}
\end{tcolorbox}

\begin{tcolorbox}[llmbox2,title=\textbf{LLM output}]
{\ttfamily\small
Okay, so I need to figure out the top factors suggesting severe chest trauma based on the given ICU free-text note. Let me start by reading through the note carefully.

First, the patient is a \#\#-year-old male who fell about 10 feet from a tree stand. He's on warfarin, which is a blood thinner, so that might contribute to bleeding issues. He has an obvious deformity on the left shoulder and is clutching the area, which could indicate a fracture or significant injury there.

Looking at the impressions, there are multiple left-sided rib fractures involving ribs 1-7 and 11. That's a lot of broken ribs, which is a strong indicator of severe chest trauma. Also, there's a mention of a left apical pneumothorax, which is a collection of air in the chest cavity, causing the lung to collapse. Additionally, there's a small left hemothorax, meaning there's blood accumulating around the lung, which can be serious.

The patient also has a comminuted fracture of the left clavicle and a nondisplaced fracture of the left scapula. These injuries, especially the clavicle fracture being comminuted (meaning it's shattered into pieces), suggest significant force was applied to the chest area.

His breathing is described as fast and shallow, which could indicate pain or difficulty due to the chest injuries. However, since the output rules specify not to mention treatment or prognosis, I should focus on the structural injuries.

So, the key factors here are the multiple rib fractures, the pneumothorax, and the hemothorax. These are all directly related to severe chest trauma. The clavicle and scapula fractures also support the severity but might be secondary to the chest injuries. The fall from height and the use of blood thinners are contributing factors but aren't direct indicators of chest trauma severity as per the note.

I think the top three factors would be the multiple rib fractures, the presence of pneumothorax, and the hemothorax. Each of these points is explicitly mentioned and strongly suggests severe chest trauma.
</think>

1. Multiple left-sided rib fractures — Indicates severe chest wall damage.  
   Evidence: "multiple left - sided rib fractures involving ribs \# 1 - 7 and 11"

2. Left apical pneumothorax — Suggests lung compression or collapse.  
   Evidence: "left apical pneumothorax"

3. Small left hemothorax — Indicates internal bleeding in the chest.  
   Evidence: "small left hemothorax"

4. Comminuted fracture of the left clavicle — Severe bony injury reflecting blunt force.  
   Evidence: "comminuted fracture left clavicle"

5. Nondisplaced fracture of the left scapula — Additional evidence of significant trauma.  
   Evidence: "nondisplaced fracture of the left scapula"
}
\end{tcolorbox}

\subsection*{E. Perturbation results}

Across both tasks and all encoder backbones, perturbing tokens associated with the most contributing concepts resulted in larger performance drops than token-count-matched random perturbation (Table \ref{tab:supp_perturbation_control}, \ref{tab:supp_perturbation_control_random_tokens}). The drop was larger when positive-class contributors were masked, and the performance drop became larger when masking the top 10 contributors rather than only the top 5.

\begin{table}[p]
\centering
\vspace{-1cm}
\caption{AUROC under concept-based perturbation and corresponding
token-count-matched control perturbation. Random tokens in the control comparisons are replaced by \texttt{[MASK]} token. Values are reported as
mean $\pm$ standard deviation across runs.}
\label{tab:supp_perturbation_control}
\small
\setlength{\tabcolsep}{6pt}

\textbf{(a) Chest trauma severity}

\medskip

\begin{tabular}{llcc}
\toprule
Backbone & Perturbation & Masked & Control \\
\midrule

\multirow{5}{*}{SapBERT}
& Blackbox        & \multicolumn{2}{c}{$0.962 \pm 0.012$} \\
& Negative Top-5  & $0.921 \pm 0.020$ & $0.947 \pm 0.016$ \\
& Negative Top-10 & $0.905 \pm 0.031$ & $0.935 \pm 0.013$ \\
& Positive Top-5  & $0.824 \pm 0.031$ & $0.920 \pm 0.018$ \\
& Positive Top-10 & $0.786 \pm 0.106$ & $0.891 \pm 0.044$ \\
\addlinespace

\multirow{5}{*}{ClinicalBERT}
& Blackbox        & \multicolumn{2}{c}{$0.959 \pm 0.011$} \\
& Negative Top-5  & $0.947 \pm 0.007$ & $0.950 \pm 0.008$ \\
& Negative Top-10 & $0.937 \pm 0.009$ & $0.941 \pm 0.011$ \\
& Positive Top-5  & $0.702 \pm 0.139$ & $0.853 \pm 0.051$ \\
& Positive Top-10 & $0.688 \pm 0.114$ & $0.774 \pm 0.096$ \\
\addlinespace

\multirow{5}{*}{BGE}
& Blackbox        & \multicolumn{2}{c}{$0.957 \pm 0.010$} \\
& Negative Top-5  & $0.898 \pm 0.047$ & $0.904 \pm 0.035$ \\
& Negative Top-10 & $0.840 \pm 0.033$ & $0.896 \pm 0.028$ \\
& Positive Top-5  & $0.786 \pm 0.013$ & $0.845 \pm 0.019$ \\
& Positive Top-10 & $0.760 \pm 0.092$ & $0.801 \pm 0.080$ \\
\bottomrule
\end{tabular}

\vspace{0.7cm}

\textbf{(b) In-hospital mortality}

\medskip

\begin{tabular}{llcc}
\toprule
Backbone & Perturbation & Masked & Control \\
\midrule

\multirow{5}{*}{SapBERT}
& Blackbox        & \multicolumn{2}{c}{$0.845 \pm 0.017$} \\
& Negative Top-5  & $0.752 \pm 0.032$ & $0.821 \pm 0.028$ \\
& Negative Top-10 & $0.720 \pm 0.063$ & $0.793 \pm 0.054$ \\
& Positive Top-5  & $0.673 \pm 0.044$ & $0.765 \pm 0.037$ \\
& Positive Top-10 & $0.611 \pm 0.018$ & $0.722 \pm 0.029$ \\
\addlinespace

\multirow{5}{*}{ClinicalBERT}
& Blackbox        & \multicolumn{2}{c}{$0.849 \pm 0.020$} \\
& Negative Top-5  & $0.754 \pm 0.033$ & $0.830 \pm 0.025$ \\
& Negative Top-10 & $0.718 \pm 0.043$ & $0.806 \pm 0.036$ \\
& Positive Top-5  & $0.692 \pm 0.015$ & $0.781 \pm 0.022$ \\
& Positive Top-10 & $0.626 \pm 0.021$ & $0.739 \pm 0.030$ \\
\addlinespace

\multirow{5}{*}{BGE}
& Blackbox        & \multicolumn{2}{c}{$0.839 \pm 0.024$} \\
& Negative Top-5  & $0.797 \pm 0.013$ & $0.825 \pm 0.027$ \\
& Negative Top-10 & $0.724 \pm 0.028$ & $0.818 \pm 0.032$ \\
& Positive Top-5  & $0.749 \pm 0.017$ & $0.778 \pm 0.025$ \\
& Positive Top-10 & $0.663 \pm 0.026$ & $0.765 \pm 0.034$ \\
\bottomrule
\end{tabular}

\end{table}

\begin{table}[p]
\centering
\vspace{-1cm}
\caption{AUROC under concept-based perturbation and corresponding
token-count-matched control perturbation. Random tokens in the control comparisons are replaced by a random token. Values are reported as mean $\pm$ standard deviation across runs.}
\label{tab:supp_perturbation_control_random_tokens}
\small
\setlength{\tabcolsep}{6pt}

\textbf{(a) Chest trauma severity}

\medskip

\begin{tabular}{llcc}
\toprule
Backbone & Perturbation & Perturbed & Control \\
\midrule

\multirow{5}{*}{SapBERT}
& Blackbox        & \multicolumn{2}{c}{$0.962 \pm 0.012$} \\
& Negative Top-5  & $0.903 \pm 0.024$ & $0.936 \pm 0.021$ \\
& Negative Top-10 & $0.882 \pm 0.034$ & $0.929 \pm 0.017$ \\
& Positive Top-5  & $0.801 \pm 0.036$ & $0.915 \pm 0.023$ \\
& Positive Top-10 & $0.758 \pm 0.111$ & $0.886 \pm 0.050$ \\
\addlinespace

\multirow{5}{*}{ClinicalBERT}
& Blackbox        & \multicolumn{2}{c}{$0.959 \pm 0.011$} \\
& Negative Top-5  & $0.932 \pm 0.010$ & $0.942 \pm 0.014$ \\
& Negative Top-10 & $0.918 \pm 0.013$ & $0.938 \pm 0.013$ \\
& Positive Top-5  & $0.679 \pm 0.145$ & $0.842 \pm 0.067$ \\
& Positive Top-10 & $0.654 \pm 0.121$ & $0.759 \pm 0.100$ \\
\addlinespace

\multirow{5}{*}{BGE}
& Blackbox        & \multicolumn{2}{c}{$0.957 \pm 0.010$} \\
& Negative Top-5  & $0.879 \pm 0.050$ & $0.903 \pm 0.045$ \\
& Negative Top-10 & $0.819 \pm 0.038$ & $0.881 \pm 0.038$ \\
& Positive Top-5  & $0.765 \pm 0.017$ & $0.836 \pm 0.025$ \\
& Positive Top-10 & $0.731 \pm 0.097$ & $0.800 \pm 0.073$ \\
\bottomrule
\end{tabular}

\vspace{0.7cm}

\textbf{(b) In-hospital mortality}

\medskip

\begin{tabular}{llcc}
\toprule
Backbone & Perturbation & Perturbed & Control \\
\midrule

\multirow{5}{*}{SapBERT}
& Blackbox        & \multicolumn{2}{c}{$0.845 \pm 0.017$} \\
& Negative Top-5  & $0.731 \pm 0.037$ & $0.813 \pm 0.031$ \\
& Negative Top-10 & $0.694 \pm 0.067$ & $0.795 \pm 0.061$ \\
& Positive Top-5  & $0.648 \pm 0.049$ & $0.736 \pm 0.045$ \\
& Positive Top-10 & $0.582 \pm 0.022$ & $0.710 \pm 0.028$ \\
\addlinespace

\multirow{5}{*}{ClinicalBERT}
& Blackbox        & \multicolumn{2}{c}{$0.849 \pm 0.020$} \\
& Negative Top-5  & $0.733 \pm 0.038$ & $0.815 \pm 0.027$ \\
& Negative Top-10 & $0.695 \pm 0.048$ & $0.810 \pm 0.039$ \\
& Positive Top-5  & $0.668 \pm 0.019$ & $0.769 \pm 0.023$ \\
& Positive Top-10 & $0.596 \pm 0.025$ & $0.741 \pm 0.032$ \\
\addlinespace

\multirow{5}{*}{BGE}
& Blackbox        & \multicolumn{2}{c}{$0.839 \pm 0.024$} \\
& Negative Top-5  & $0.779 \pm 0.016$ & $0.822 \pm 0.028$ \\
& Negative Top-10 & $0.703 \pm 0.033$ & $0.820 \pm 0.042$ \\
& Positive Top-5  & $0.724 \pm 0.020$ & $0.741 \pm 0.029$ \\
& Positive Top-10 & $0.635 \pm 0.029$ & $0.758 \pm 0.044$ \\
\bottomrule
\end{tabular}

\end{table}

\newpage
\subsection*{F. Performance without warmup}

\begin{table}[ht]
\centering
\caption{Performance comparison between blackbox models and our framework with and without the warmup described in \ref{method:encoder}. Adding the warmup substantially narrows the performance gap, enabling our framework to achieve performance comparable to the corresponding blackbox models.}

\resizebox{\textwidth}{!}{
\begin{tabular}{*5l}
\toprule
\label{tab:supp_warmup_ablation}
{} & \multicolumn{2}{c}{Chest trauma severity} & \multicolumn{2}{c}{In-hospital mortality} \\
{} & AUROC & AUPR & AUROC & AUPR \\
\midrule
SapBERT blackbox & \best{$0.962\,{\scriptstyle \pm\,0.012}$} & \best{$0.873\,{\scriptstyle \pm\,0.010}$} & \best{$0.845\,{\scriptstyle \pm\,0.017}$} & \best{$0.346\,{\scriptstyle \pm\,0.015}$} \\
Without warmup (SapBERT backbone) & $0.803\,{\scriptstyle \pm\,0.021}$ & $0.658\,{\scriptstyle \pm\,0.036}$ & $0.657\,{\scriptstyle \pm\,0.032}$ & $0.102\,{\scriptstyle \pm\,0.026}$ \\
Ours (SapBERT backbone) & $0.959\,{\scriptstyle \pm\,0.013}$\nearbb & \best{$0.873\,{\scriptstyle \pm\,0.009}$}\nearbb & $0.839\,{\scriptstyle \pm\,0.016}$\nearbb & $0.332\,{\scriptstyle \pm\,0.018}$\nearbb \\
\midrule
ClinicalBERT & \best{$0.959\,{\scriptstyle \pm\,0.011}$} & \best{$0.871\,{\scriptstyle \pm\,0.010}$} & \best{$0.849\,{\scriptstyle \pm\,0.020}$} & \best{$0.325\,{\scriptstyle \pm\,0.017}$} \\
Without warmup (ClinicalBERT backbone) & $0.790\,{\scriptstyle \pm\,0.033}$ & $0.652\,{\scriptstyle \pm\,0.029}$ & $0.701\,{\scriptstyle \pm\,0.037}$ & $0.115\,{\scriptstyle \pm\,0.019}$ \\
Ours (ClinicalBERT backbone) & $0.957\,{\scriptstyle \pm\,0.012}$\nearbb & $0.864\,{\scriptstyle \pm\,0.011}$\nearbb & $0.833\,{\scriptstyle \pm\,0.022}$\nearbb & $0.314\,{\scriptstyle \pm\,0.019}$\nearbb \\
\midrule
BGE & \best{$0.957\,{\scriptstyle \pm\,0.010}$} & \best{$0.861\,{\scriptstyle \pm\,0.012}$} & $0.839\,{\scriptstyle \pm\,0.024}$ & $0.309\,{\scriptstyle \pm\,0.015}$ \\
Without warmup (BGE backbone) & $0.756\,{\scriptstyle \pm\,0.033}$ & $0.527\,{\scriptstyle \pm\,0.036}$ & $0.658\,{\scriptstyle \pm\,0.019}$ & $0.121\,{\scriptstyle \pm\,0.015}$ \\
Ours (BGE backbone) & $0.937\,{\scriptstyle \pm\,0.014}$\nearbb & $0.821\,{\scriptstyle \pm\,0.013}$ & \best{$0.843\,{\scriptstyle \pm\,0.016}$}\nearbb & \best{$0.318\,{\scriptstyle \pm\,0.018}$}\nearbb \\
\bottomrule
\end{tabular}
}
\end{table}

\subsection*{G. Statistics of masked tokens in perturbation study} 

\begin{table}[htbp]
\centering
\caption{Number and percentage of tokens removed in the concept perturbation analysis. Values are reported as mean (SD) number of tokens / mean (SD) percentage of tokens removed. }
\label{tab:supp_masked_tokens}
\scriptsize
\setlength{\tabcolsep}{3.5pt}
\begin{tabular}{llcccc}
\toprule
& & \multicolumn{2}{c}{Chest trauma} &
     \multicolumn{2}{c}{In-hospital mortality} \\
% \cmidrule(lr){3-4}\cmidrule(lr){5-6}
\mycmidrule{3-4}\mycmidrule{5-6}

Backbone & Top-$k$
& \shortstack{Positive-class\\concepts}
& \shortstack{Negative-class\\concepts}
& \shortstack{Positive-class\\concepts}
& \shortstack{Negative-class\\concepts} \\
% \cmidrule(lr){3-6}
\mycmidrule{3-6}
& & \multicolumn{4}{c}{\textit{Tokens removed: mean number (SD) / mean percentage (SD \%)}} \\
\midrule
SapBERT
 & 5  & 85 (37) / 21 (10)\% & 22 (9) / 5 (2)\%
      & 103 (41) / 25 (12)\% & 30 (9) / 8 (3)\% \\
 & 10 & 127 (46) / 32 (11)\% & 34 (10) / 8 (3)\%
      & 120 (54) / 32 (15)\% & 40 (13) / 10 (4)\% \\
\addlinespace
ClinicalBERT
 & 5  & 101 (46) / 23 (12)\% & 29 (11) / 5 (2)\%
      & 114 (57) / 28 (13)\% & 36 (10) / 8 (3)\% \\
 & 10 & 143 (58) / 36 (12)\% & 49 (11) / 11 (3)\%
      & 152 (69) / 37 (17)\% & 46 (15) / 11 (4)\% \\
\addlinespace
BGE
 & 5  & 77 (39) / 17 (9)\% & 23 (12) / 2 (1)\%
      & 89 (42) / 19 (9)\% & 33 (10) / 6 (2)\% \\
 & 10 & 105 (25) / 26 (8)\% & 31 (10) / 8 (3)\%
      & 99 (42) / 25 (10)\% & 38 (10) / 10 (3)\% \\
\bottomrule
\end{tabular}
\end{table}

\end{document}